\documentclass{article}

\usepackage{arxiv}

\usepackage[utf8]{inputenc} 
\usepackage[T1]{fontenc}    
\usepackage{hyperref}       
\usepackage{url}            
\usepackage{booktabs}       
\usepackage{amsfonts}       
\usepackage{nicefrac}       
\usepackage{microtype}      
\usepackage{lipsum}         
\usepackage{graphicx}
\usepackage{natbib}
\usepackage{doi}

\usepackage[table]{xcolor}
\usepackage{algorithm}
\usepackage{algorithmic}

\usepackage{amsmath,amsfonts,bm}

\def\eqref#1{equation~\ref{#1}}

\def\1{\bm{1}}

\def\vg{{\bm{g}}}

\def\vx{{\bm{x}}}

\def\vz{{\bm{z}}}

\DeclareMathAlphabet{\mathsfit}{\encodingdefault}{\sfdefault}{m}{sl}
\SetMathAlphabet{\mathsfit}{bold}{\encodingdefault}{\sfdefault}{bx}{n}

\newcommand{\todo}[1]{}

\definecolor{Gray}{gray}{0.9}
\definecolor{citecolor}{HTML}{2980b9}
\definecolor{linkcolor}{HTML}{c0392b}

\usepackage{longtable}

\usepackage{subcaption}
\usepackage{mathrsfs}

\usepackage{bm}

\newlength\savewidth\newcommand\shline{\noalign{\global\savewidth\arrayrulewidth
  \global\arrayrulewidth 1pt}\hline\noalign{\global\arrayrulewidth\savewidth}}
\newcommand{\tablestyle}[2]{\setlength{\tabcolsep}{#1}\renewcommand{\arraystretch}{#2}\centering\footnotesize}
\renewcommand{\paragraph}[1]{\vspace{1.25mm}\noindent\textbf{#1}}

\newcolumntype{x}[1]{>{\centering\arraybackslash}p{#1pt}}
\newcolumntype{y}[1]{>{\raggedright\arraybackslash}p{#1pt}}
\newcolumntype{z}[1]{>{\raggedleft\arraybackslash}p{#1pt}}

\definecolor{deemph}{gray}{0.6}

\definecolor{baselinecolor}{gray}{.9}
\newcommand{\baseline}[1]{\cellcolor{baselinecolor}{#1}}
\newcommand{\default}{\rowcolor{gray!30}}

\usepackage{wrapfig}
\usepackage{enumitem}

\newcommand{\ours}{{SimLAP}}

\usepackage{cleveref}
\crefformat{section}{Section.~#2#1#3}
\crefformat{subsection}{Section.~#2#1#3}
\crefformat{subsubsection}{Sec.~#2#1#3}
\crefformat{figure}{Figure.~#2#1#3}
\crefformat{equation}{Equation.~#2#1#3}
\crefformat{table}{Table.~#2#1#3}
\crefformat{algorithm}{Alg.~#2#1#3}

\title{
Rethinking Positive Pairs in Contrastive Learning
}

\usepackage{authblk}

\author[1,2]{Jiantao Wu}
\author[1,2]{Sara Atito}
\author[4]{Zhenhua Feng}
\author[3]{Shentong Mo}
\author[2]{Josef Kitler}
\author[1,2]{Muhammad Awais}
\affil[1]{Surrey Institute for People-Centred AI, GU2 7XH Surrey, UK}
\affil[2]{Centre For Vision, Speech and Signal Processing (CVSSP), GU2 7XH Surrey, UK}
\affil[3]{Carnegie Mellon University, 5000 Forbes Ave, Pittsburgh, PA 15213, Pennsylvania, USA}
\affil[4]{Jiangnan University, 1800 Lihu Avenue, Wuxi, Jiangsu, P. R. China}

\renewcommand{\shorttitle}{\ours{}}

\hypersetup{
pdftitle={A template for the arxiv style},
pdfsubject={unpublished},
pdfauthor={J.T.},
pdfkeywords={paper},
}

\begin{document}

\maketitle


\begin{abstract}

Contrastive learning, a prominent approach to representation learning, traditionally assumes positive pairs are closely related samples (the same image or class) and negative pairs are distinct samples.  We challenge this assumption by proposing to learn from arbitrary pairs, allowing any pair of samples to be positive within our framework.
The primary challenge of the proposed approach lies in applying contrastive learning to disparate pairs which are semantically distant. 
Motivated by the discovery that SimCLR can separate given arbitrary pairs (e.g., garter snake and table lamp) in a subspace, we propose a feature filter in the condition of class pairs that creates the requisite subspaces by gate vectors selectively activating or deactivating dimensions. This filter can be optimized through gradient descent within a conventional contrastive learning mechanism.

We present \ours{}, a universal contrastive learning framework for visual representations that extends conventional contrastive learning to accommodate arbitrary pairs. 
Our approach is validated using IN1K, where 1K diverse classes compose 500,500 pairs, most of them being distinct. Surprisingly, \ours{} achieves superior performance in this challenging setting. Additional benefits include the prevention of dimensional collapse and the discovery of class relationships. 
\todo{insights:}
Our work highlights the value of learning common features of arbitrary pairs and potentially broadens the applicability of contrastive learning techniques on the sample pairs with weak relationships.
\end{abstract}

\section{Introduction}

Contrastive learning (CL) has demonstrated its efficacy in the domain of visual representation, substantially enhancing the state-of-the-art outcomes across a spectrum of visual tasks, as evidenced by recent studies \citep{ chen2020simple,he2020momentum}. 
The fundamental principle of contrastive learning is to cultivate the emergence of discriminative features that enable the differentiation between positive and negative samples. 
Conventionally, CL assumes that positive samples share more common features than negatives.
In the context of self-supervised or instance-wise contrastive learning (ICL), this is typically achieved through data augmentation, such as random cropping and color variation~\citep{chen2020simple}, which preserve semantic information while introducing perturbations to prevent trivial solutions. 
In class-wise contrastive learning (CCL)~\citep{khosla2020supervised,cui2021parametric}, the positive pairs are from the same class sharing class-relevant features and, thus, are closer than those from different classes.
Such an idea is natural and intuitive, as it is widely accepted that positive pairs should have a close relationship. We pose the question: \textit{is it possible to compose positive pairs from two disparate images belonging to different classes, e.g., snake and cat?} We consider negative samples are not positive in the same batch and rethink the positive pairs in CL from the following scenarios:

\begin{figure}[tbh]
    \centering
    \begin{subfigure}{0.15\linewidth}
        \includegraphics[width=.92\linewidth]{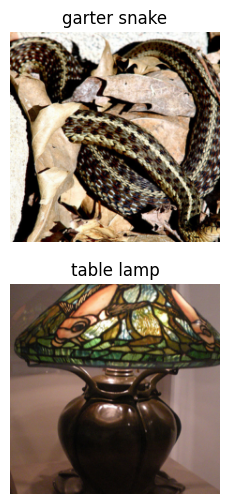}%
        \subcaption[]{Example}
    \end{subfigure}
    \begin{subfigure}{0.8\linewidth}
    \centering
        \includegraphics[width=.9\linewidth]{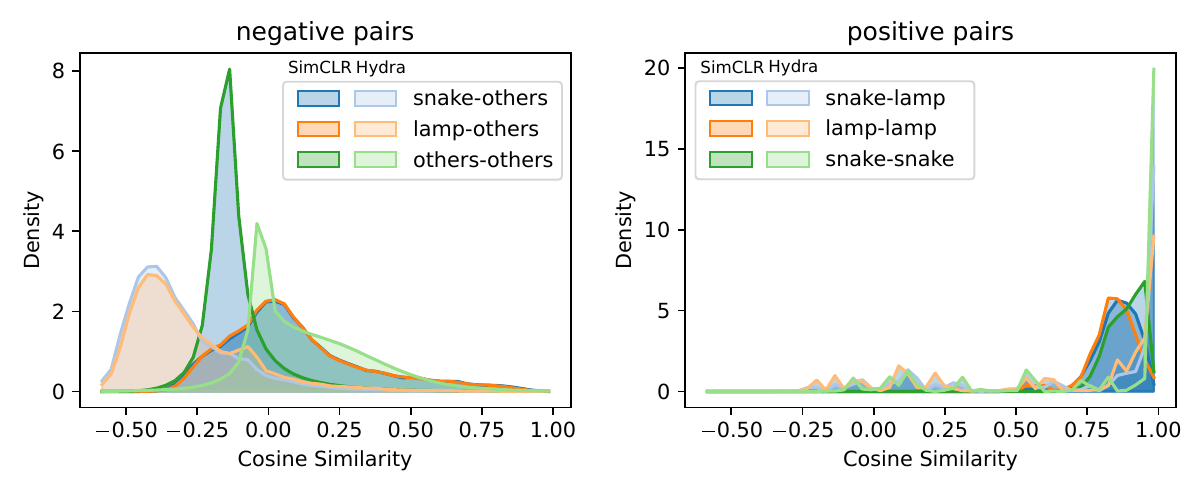}%
        \subcaption[]{Subspace for snake-lamp}
        \label{fig:dist_snake_lamp}
    \end{subfigure}
    \caption{\textbf{Similarity distribution of class pairs under the subspaces} of feature extracted by SimCLR and \ours{} for snake-lamp.
    We can hardly find any common visual features from the example of a garter snake and a table lamp, However, we find that the 500 dimensions with the lowest variance from SimCLR's representation can separate snake-lamp from other classes. \ours{} learns such a subspace while representation learning.}
    \label{fig:intuition}
\end{figure}

\paragraph{(i) Similar samples.} 
When the positive pairs are from the same sample or the same category, they have closely related semantic content and typically share a substantial number of characteristic features. In contrast, the  negative samples are randomly sampled, sharing a few or no features. The CL methods can extract these discriminative features can easily separate them from negative samples.

\paragraph{(ii) Disparate samples.}
The semantic distance between disparate samples may be far away than two random samples (negative pairs). Therefore, disparate samples share less features than negatives. Here, dissimilarities outweigh similarities, making their representations less discriminative. Therefore, conventional CL methods struggle with such disparate positive pairs and only work with similar pairs~\citep{chen2020simple,weinberger2009distance,oord2018representation}.
However, the common features exist among any class pair and are discriminative to differentiate them from other classes, even they are inscrutable, not meaningful, and not valuable. 
\cref{fig:intuition} validates this hypothesis by demonstrating a subspace for garter snake and lamp pair, where 500 out of 2048 dimensions in SimCLR's representation differentiate this pair from other classes.


This observation motivates our development of universal contrastive learning (UCL), allowing arbitrary pairs to be positive by creating a subspace for each class pair.
We propose a learning principle for UCL that creates foundations for learning visual representations from arbitrary samples while remaining robust to disparate samples. This approach addresses two key challenges:
1) Selectively promoting discriminatory information in subspaces while avoiding penalization of irrelevant features in disparate pairs.
2) Identifying subspaces for arbitrary pairs with limited supervision.
To overcome these challenges, we introduce a novel ``feature filter'' module. This filter utilizes the class pair to identify subspaces and outputs gates with weights ranging from 0 to 1, effectively creating a unique subspace for each pair. By confining the impact of contrastive learning loss to a subspace, our approach prevents erroneous contrast behavior from affecting representation learning in other spaces.


We present \ours{} by extending CL methods to accommodate arbitrary pairs, e.g., SimCLR. 
To show the robustness of dealing with disparate pairs, we train \ours{} on ImageNet~\citep{deng2009imagenet}, with 1,000 diverse classes forming 500,500 class pairs, most of which are disparate. The positive pairs are randomly selected from the 500,500 class pairs. 
Our model achieves superior performance in the challenging situation, demonstrating its robustness and effectiveness in discovering common features.
\todo{contribution and insights from this work}
In summary, our contributions are as follows: 
\begin{itemize}[noitemsep]
    \item We propose \ours{} a universal contrastive learning framework to allow learning visual representations from samples belonging to arbitrary class pairs through feature selection.
    \item We show that learning common features from arbitrary pairs prevents dimensional collapse and promotes transferable representations.
    \item Our framework is able to uncover the class specific properties  of similar samples, as well as  to learn from disparate samples.
    \item Our work expands the range of positive pairs to arbitrary pairs, potentially broadening the applicability of contrastive learning techniques.
\end{itemize}


\section{Related Work}

\paragraph{Contrastive Learning.}
Contrastive learning has emerged as a powerful self-supervised learning (SSL) paradigm for learning effective visual representations from unlabeled data, which enables transfer learning~\citep{he2020momentum}. The fundamental principle of contrastive learning is to promote invariant semantics from positive pairs by minimizing the distance between similar instances while maximizing the distance between dissimilar instances in the embedding space.
Early work in this area introduced the triplet loss~\citep{weinberger2009distance,schroff2015facenet}, which aims to learn an embedding space where positive pairs are pulled together while negative pairs are pushed apart by a certain margin. Subsequently, \cite{wu2018unsupervised} proposed Instance Discrimination, treating each instance as a distinct category and training a classifier to identify it among a set of negative instances.
InfoNCE~\citep{oord2018representation} is a loss function that maximizes mutual information between the encoded representations of different views of the same data point. 
SimCLR~\citep{chen2020simple} uses a composition of data augmentations and a contrastive loss to learn representations by maximizing the agreement between differently augmented views of the same image (positive pairs) while pushing the other images in one batch (negative pairs) away. 
A dynamic dictionary with a queue and a moving-averaged encoder has been proposed to provide stable tokens to enhance performance~\citep{he2020momentum}.

\paragraph{Positive Pair Design.}
The design of positive pairs is crucial in contrastive learning, as it directly impacts the quality of learned representations. Recent work has focused on expanding the scope of what constitutes a positive pair. 
Instance-wise Contrastive Learning (ICL): positive pairs typically come from the same instance but use hand-crafted data augmentation ~\citep{oord2018representation,Grill20_BYOL,caron2020unsupervised}. Data augmentation is crucial to prevent the model learning trivial representations and improve the transfer learning performance~\citep{chen2020simple}. Class-wise Contrastive Learning (CCL): CCL methods~\citep{cui2021parametric,cui2023generalized} extend the range of positive pairs to samples from the same class.Contrastive Learning with Auxiliary-Information: Constructing positive pairs based on specific variables, such as auxiliary information~\citep{tsai2021integrating} and attributes~\citep{ma2021conditional}. Nevertheless, the above methods highly rely on a prior to construct positive pairs.

Recent work has begun to explore more complex relationships between samples. Positive Active Learning (PAL)~\citep{Cabannes_2023_ICCV} selects positive pairs based on a similarity graph. $\mathbb{X}$-CLR~\citep{sobal2024_Xsample} utilises text caption to facilitate the calculation of the similarity graph. 
These methods are still limited to optimizing closely related samples and do not address the challenge of learning from disparate samples, like garter snake and lamp. 

Our work not only widens the scope of positive pairs beyond closely related samples but also reveals that similarity graphs should be high-dimensional instead of 2D to denote the complex structure of relationships between samples. This insight allows for a more nuanced representation of sample similarities, capturing subtle relationships that might be lost in one global feature space.

\paragraph{Dimensional Collapse.}
Though CL methods are capable to avoid a complete collapse, they still suffer from dimensional collapse~\citep{hua2021feature}, where the representations are collapsed to a lower-dimensional space. SimCLR attaches a non-linear projector after the backbone to overcome the dimensional collapse~\citep{DBLP:conf/iclr/JingVLT22}.
However, empirical evidence suggests that this only partially alleviates the issue, especially in long-term training scenarios. 
Our framework addresses the dimensional collapse more effectively by learning from arbitrary pairs, which contain diverse features. 
\cite{zhang2024avoiding}

\section{Method}

\subsection{Intuition}

Our approach is founded on the hypothesis that seemingly disparate classes share common features that may not be immediately apparent to human observers. We discovered that common features can be found among disparate classes. 

We consider the seemingly unrelated classes of ``garter snake'' and ``table lamp'' to be positive. We extract features from the IN1K validation set using a pretrained SimCLR. 
Although it is inscrutable for humans to find any common features among snake and lamp\footnote{The answer from Claude: They both have a long, slender body with a wider base, smooth surfaces, and a distinct ``head'' at one end, though the snake's is more flexible and the lamp's houses a light bulb.}, we can identify a subspace for the snake-lamp by selecting 500 dimensions with the lowest variance for the snake-lamp pair for SimCLR, which likely represent common features. 
We draw the similarity distribution under the subspaces for negative pairs and positive pairs in \cref{fig:dist_snake_lamp}. 
SimCLR finds the common features for snake-lamp having a high intra similarity and a low inter similarity!


This finding underscores the potential of universal contrastive learning by creating subspaces to extract common features for arbitrary pairs, forming the foundation of our approach. By leveraging these implicit commonalities, the model can exploit the shared features to improve the quality of the learned representations. Motivated by this, we develop a framework to extend CL methods for arbitrary pairs by learning the corresponding subspaces.

\subsection{Contrastive Learning}

We revisit the basics of contrastive learning, which promotes the discovery of discriminative features between positive and negative samples. The key to contrastive learning is defining positives and negatives. For each anchor sample in the dataset $\mathbf{x} \in \mathcal{X}$, a conventional contrastive learning method defines its positives $\mathbf{x}^+$ and negatives $\mathbf{x}^-$ according to its needs.
An encoder network $f_V(\cdot)$ is applied to map the images to a representation vector. A projection network $f_P$ is critical to mitigating a dimensional collapse~\citep{DBLP:conf/iclr/JingVLT22} and to improving the transfer learning performance~\citep{chen2020simple}. Overall, the features are extracted by $\vz=f_P(f_V(\bm{x}))$.
The objective helps to identify the features that can separate the positives and negatives. InfoNCE~\citep{oord2018representation} is widely applied to achieve this:
\begin{equation}
    \mathcal{L} = - \frac{ \exp(\mathrm{sim}((\vz), (\vz^+))) / \tau}{\exp(\mathrm{sim}((\vz), (\vz^+))) / \tau +   \sum_{\vz^- \in \mathbf{\bm{Z}^-}}{ \exp(\mathrm{sim}((\vz), (\vz^-)) / \tau) }},
\end{equation}
where $\mathrm{sim}(u,v)=\frac{u v}{\|u\| \|v\|}$ denotes the cosine similarity between two vectors.

We argue that positive samples do not have to be limited to closely related samples. They can be arbitrary pairs, including different views of the same image, different samples from the same class, or disparate samples from different classes, as long as there is discriminative information.

\subsection{Universal Contrastive Learning}

\begin{figure}
    \centering
    \includegraphics[width=0.7\linewidth]{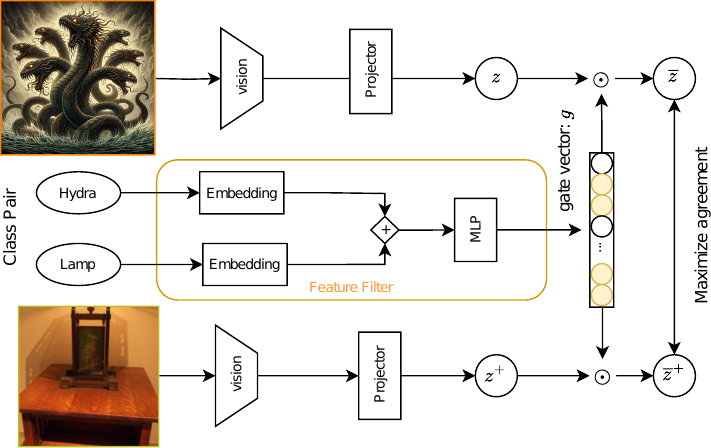}
    \caption{\textbf{Universal contrastive learning for arbitrary class pairs}. The feature filter generates a gate vector to activate the common features for the given class pair by averaging their label embeddings. \ours{} learns visual representations by maximizing the agreement of common features between disparate samples (Hydra-Lamp) in the corresponding subspace.}
    \label{fig:method}
\end{figure}

In principle, our framework is adaptable to most SSL methods to allow arbitrary pairs to be positive. For simplicity, our discussion is based on SimCLR and extends it by inserting a feature filter module as illustrated in \cref{fig:method}. 
Specifically, the feature filter selectively activates certain dimensions, effectively generating a subspace for a class pair to represent their common features. This approach enables the discovery and utilization of shared information between seemingly disparate samples, expanding the scope of positive pairs in contrastive learning. We randomly choose one label from a mini-batch as $y2$ and the label of anchor sample $y1$, thus, the samples belonging to $y2$ are positive, and the samples not belonging to $y1$ nor $y2$ are negative.
The loss function for our universal contrastive learning framework is formulated as follows:
\begin{equation}
     \mathcal{L}(y1,y2) = -  \log \frac{ \exp(\mathrm{sim}(\bar{\vz}, \bar{\vz}^+)/ \tau) }{  \exp(\mathrm{sim}(\bar{\vz}, \bar{\vz}^+)/ \tau) 
 +\sum_{\bar{\vz}^- \notin y1,y2}{ \exp(\mathrm{sim}(\bar{\vz}, \bar{\vz}^-) / \tau) }},
\end{equation}
where $\{\bar{\vz},\bar{\vz}^+,\bar{\vz}^-\} = \vg(y1,y2) \odot \{\vz,\vz^+,\vz^-\}$ denote the features in the subspaces. The gates $\vg(y1,y2)$ control the activation of features. In this way, the InfoNCE loss works in subspaces when the gate values are binary.

Note that, the labels in our framework are utilized to select features instead of assigning samples to one corresponding cluster for each class, like supervised learning~\citep{liu2017sphereface}. Hierarchical softmax~\citep{morin2005hierarchical} and hierarchical clustering~\citep{murtagh2012hierarchical} assign samples to several semantically close clusters, but samples will be assigned to distinct clusters in our framework.

\paragraph{Feature Filter.}
The feature filter is a vital component of \ours{} to allow universal contrast learning. It controls the activation of each feature for a class pair y1-y2. The feature filter consists of two parts: label embedding and MLP layer.
Label embedding converts discrete labels to continuous vectors. We use the mean of the two label vectors to represent the common information among a class pair.
An MLP layer generates gate values to select the corresponding dimensions for the common information.
The calculation is 
\begin{align}
    \vg(y1,y2) = \sigma( f_g((f_l(y1)+f_l(y2))/2)),
\end{align}
where $f_l(\cdot)$ denotes an embedding layer to convert a label to a vector with 512 dimensions, $f_g(\cdot)$ denotes a 3-layer MLP layer, $\sigma(\cdot)$ is Sigmoid function. Note, the gates for one class are obtained when y1 equals to y2 :
\begin{equation}
    \vg(y) =  \sigma( f_g(f_l(y)) ).
\end{equation}

The flexibility of our framework allows it to potentially incorporate various types of auxiliary information to identify subspaces for arbitrary samples, such as text captions or attributes. In this paper, we focus on utilizing label information to discover common similarities between arbitrary samples. This choice is driven by our aim to prove the robustness of our framework in exploiting weak auxiliary information to infer common features and to demonstrate its effectiveness in a challenging scenario where most of the pairs are disparate.

\paragraph{Gate Penalty.}
We introduce a regularization term to promote binary values for the gates:
\begin{equation}
    \mathcal{L}_\mathcal{G} = \sum_{y=1}^K \sum_{i=1}^D \frac{1}{K D} \vg(y)_i \log{ \vg(y)}_i,
\end{equation}
where $D$ is the number of dimensions for the global feature space.
By reducing $\mathcal{L}_\mathcal{G}$, the redundancy of the representation diminishes, leading to a better interpretability. This penalty encourages the gates to be more binary, effectively selecting or deselecting features for each class pair.

\section{Experiments}

\subsection{Experimental Setting}
All experiments are conducted using ResNet50~\citep{he2016deep} as the backbone architecture, with ImageNet-1K (IN1K) as the training dataset. For baseline comparisons, we use pretrained models from official repositories: MoCov3, DINO, BarlowTwins, BYOL, and GPaCo; SimCLR implementation from VISSL~\citep{goyal2021vissl}; Supcon trained for 600 epochs. Our proposed method is implemented in two variants: \ours{} is built upon SimCLR framework, Hydra-MoCo is built upon SimCLR framework MoCov3. Detailed training configurations and hyperparameters are provided in \cref{sec:exp}.

\subsection{Main Result}

\paragraph{Transfer Learning.}
We adopt the transfer learning performance with linear prob to evaluate the quality of learned representations.
Table~\ref{tab:tl} compares UCL to ICL and CCL in terms of their transfer learning performance on six classification tasks: Cars, CIFAR10, CIFAR100, FLW, Pets, and STL, with the average (AVG) score also provided. For a fair comparison, we obtain two groups of results by using momentum or not, and pick their representatives: SimCLR as ICL and Supcon as CCL without momentum, MoCov3 as ICL and GPaCo as CCL with momentum. 
\ours{} demonstrate competitive performance in each group. Especially, \ours{} MoCo as well as MoCoV3 stands out with the highest average scores of 94.0\%, indicating their superior performance across all tasks. Notablely, \ours{} MoCo is trained for 300 epochs, using only 30\% of training for MoCov3.
Overall, the table underscores the importance of uCL in achieving optimal transfer learning performance.
\begin{table}[!thb]
\centering
\begin{tabular}{llrrrrrrr}
        & Epoch & Cars          & CF10       & CF100      & FLW           & Pets          & STL           & AVG           \\
\shline
SimCLR      & 1000  & 92.7          & \textbf{98.2}          & \textbf{85.5}          & \textbf{92.2}          & 90.8          & 97.5          & \textbf{92.8}          \\ 
Supcon      & 600   & \textbf{93.0}          & 97.4            & 85.0          & 91.1          & \textbf{92.1}          & \textbf{97.7}          & 92.7    \\
\ours{}       & 500   & 92.9          & 98.0          & 85.0          & 91.9          & 91.7          & 97.0          & 92.7          \\
\hline
\textit{momentum}    &       &               &               &               &               &               &               &               \\
MoCov3      & 1000  & 93.2          & \textbf{98.4} & 86.1          & 95.6          & \textbf{93.0} & 97.8          & \textbf{94.0} \\
GPaCo       & 700   & 92.8          & 98.3          & \textbf{86.2} & 94.1          & 92.2          & 97.9          & 93.6          \\
\ours{}-MoCo & 300   & \textbf{93.6} & 98.3          & 86.0          & \textbf{95.7} & 92.8          & \textbf{97.9} & \textbf{94.0}\\
\end{tabular}
\caption{Comparison of \textbf{transfer learning performance} of our disparate learning approach with CL methods across 6 classification tasks. The backbone is ResNet50. }\label{tab:tl}
\end{table}



\paragraph{Scaling with ViTs.}
Vision transformers~\citep{dosovitskiy2020image} are dominant in many CV tasks, especially beneficial from scaling in larger data~\citep{dehghani2023scaling}. We study the effectiveness of our model for increasing the model size in \cref{tab:scaling}. We train 100 epochs for ViT-tiny, 300 epochs for ViT-small and ViT-base. \ours{} outperforms both supervised and unsupervised learning methods  for the ViT-tiny model. The experimental results demonstrate that \ours{} scales well with increasing model size, outperforming or matching supervised learning across different ViT architectures. This scaling behavior, combined with Hydra's ability to learn from arbitrary pairs, makes it a promising approach for leveraging large-scale, diverse datasets in visual representation learning tasks.

\begin{table}[hbt]
\centering
\begin{tabular}{llll}
            &ViT-tiny & ViT-small & ViT-base \\
           \shline
Supervised  &76.94  & 92.38      &    93.49     \\
SimCLR      &69.37  & -          &  -           \\
Hydra       &\textbf{82.56}    & 91.2 & -  \\
\end{tabular}
\caption{\textbf{Scaling with ViTs.} We report top-1 accuracy on CIFAR10 with KNN. \todo{on running} }\label{tab:scaling}
\end{table}

\subsection{Universal Contrast Avoids Dimensional Collapse}
\begin{figure}[htb]
    \centering
    \begin{subfigure}{0.36\linewidth}
        \includegraphics[width=.95\linewidth]{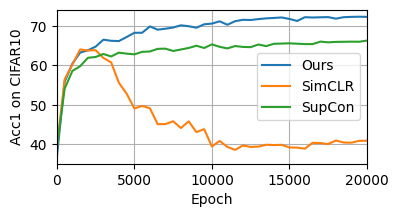}%
        \subcaption[]{Performance on CIFAR10}\label{fig:in1p_acc}
    \end{subfigure}
    \begin{subfigure}{0.62\linewidth}
        \includegraphics[width=\linewidth]{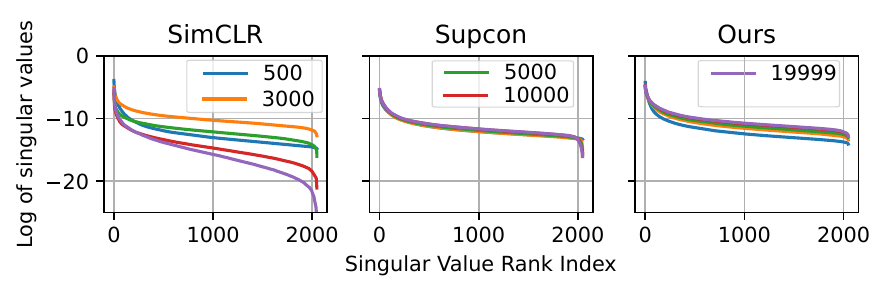}%
        \subcaption[]{Singular value spectrum of the embedding space}\label{fig:in1p_spectrum}
    \end{subfigure}
    \caption{\textbf{\ours{} prevents dimensional collapse and benefits from longer training}. Trained on IN1P and evaluated on CIFAR10. Lower singular values suggest that the learned representations are concentrating information in fewer dimensions, that is, dimensional collapse.}
    \label{fig:IN1P}
\end{figure}

To thoroughly investigate the dimensional collapse phenomenon, we construct IN1P, a focused subset of IN1K containing 12K images from 10 highly-related dog breeds. This smaller dataset allows us to extensively train on all possible class pairs, which is challenging in IN1K. We train models for 20,000 epochs to ensure comprehensive learning of all pairs. 
As shown in \cref{fig:in1p_acc}, we compare the KNN classification performance (top-1 accuracy) on CIFAR10 across different methods. SimCLR exhibits clear signs of dimensional collapse after 3,000 epochs, evidenced by declining performance and decreasing singular values~\citep{DBLP:conf/iclr/JingVLT22}. This reveals that while the non-linear projector can delay dimensional collapse, it cannot prevent it entirely. 
In contrast, \ours{} demonstrates sustained performance improvement throughout the extended training period, surpassing both Supcon and SimCLR. This robustness against dimensional collapse can be attributed to our framework's ability to leverage rich common features between classes.

\subsection{Visualization}
\begin{figure}[tbh]
    \centering
    \includegraphics[width=0.5\linewidth]{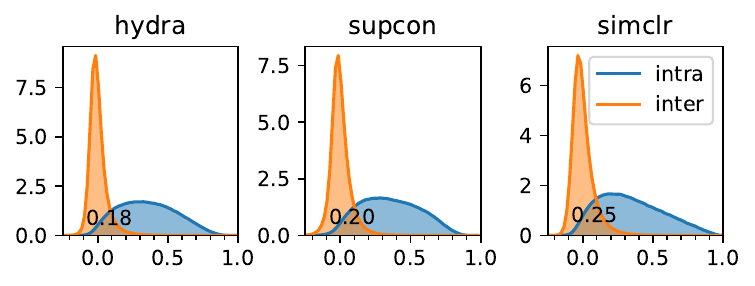}
    \caption{\textbf{Similarity distribution} for \ours{}, Supcon, and SimCLR. The number in the middle denotes the overlap of two distributions. Smaller value means better class-separation. }
    \label{fig:density_comparison}
\end{figure}

\paragraph{Embedding Analysis.}
We analyze the quality of learned representations by examining embeddings from the backbone network (excluding the projector and filter). First, we compare the inter- and intra-class similarity distributions on the IN1K validation set. As shown in \cref{fig:density_comparison}, \ours{} achieves the sharpest inter-class similarity distribution and the smallest overlap between inter- and intra-class distributions compared to Supcon and SimCLR. This indicates that \ours{} learns representations that better preserve class-specific information while maintaining clear class boundaries.

\begin{figure}[tbh]
    \centering
    \begin{subfigure}{0.24\linewidth}
        \includegraphics[width=.95\linewidth]{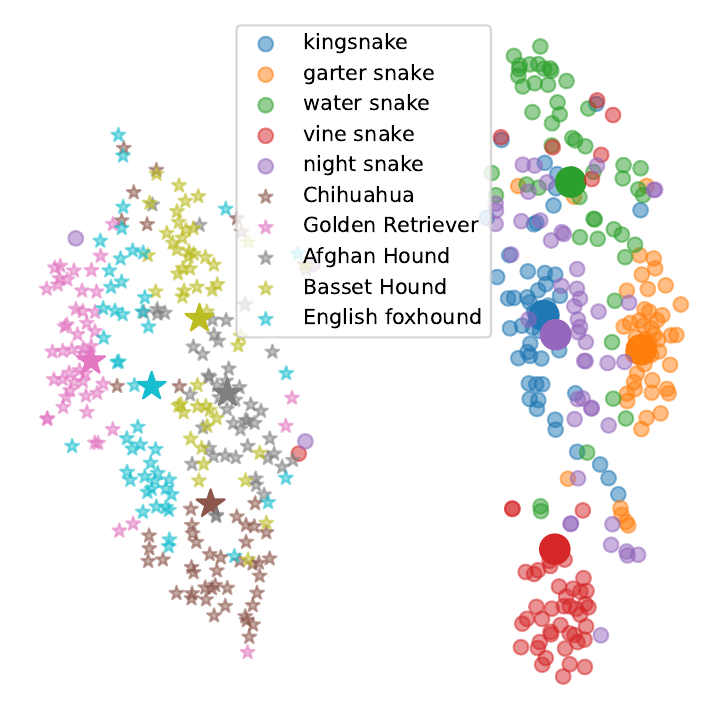}
        \subcaption[]{SimCLR}
    \end{subfigure}
    \begin{subfigure}{0.24\linewidth}
        \includegraphics[width=.95\linewidth]{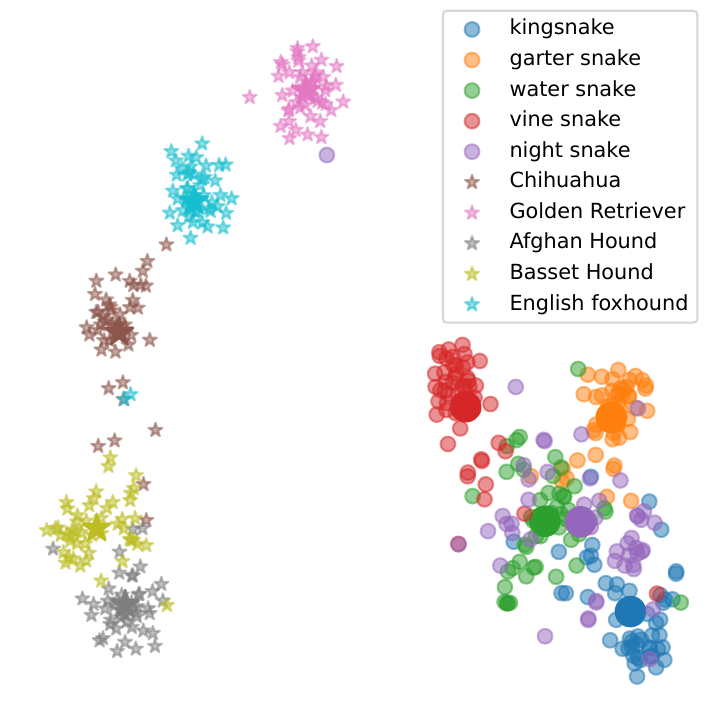}
        \subcaption[]{Supcon}
    \end{subfigure}
    \begin{subfigure}{0.24\linewidth}
        \includegraphics[width=.95\linewidth]{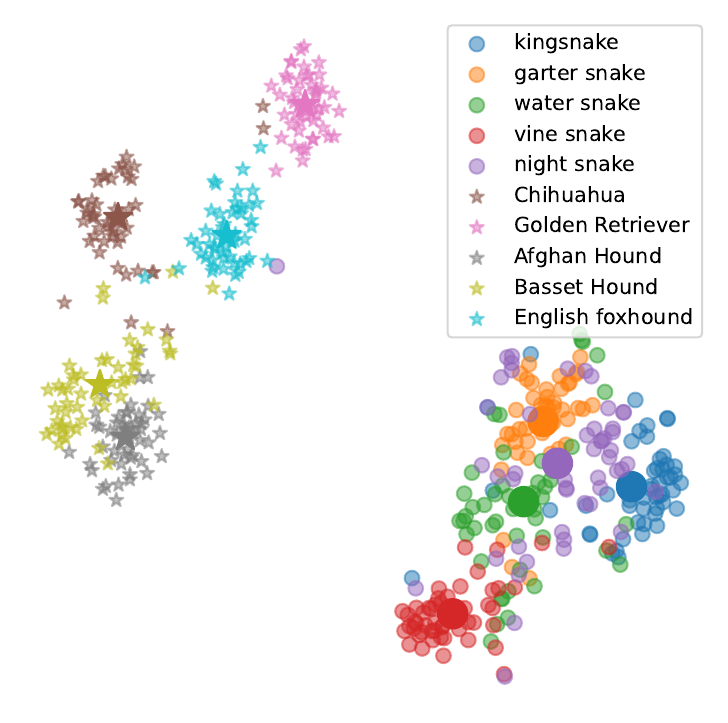}
        \subcaption[]{Hydra}
    \end{subfigure}
    \begin{subfigure}{0.24\linewidth}
        \includegraphics[width=.95\linewidth]{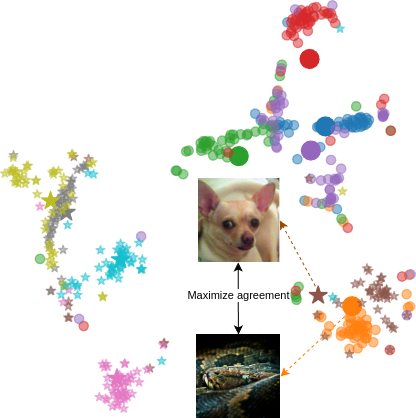}
        \subcaption[]{Hydra in subspace}
    \end{subfigure}
    \caption{\textbf{$t$-SNE visualization} of learned embeddings for two contrasting groups: 5 dog breeds and 5 snake species.  Large markers indicate class centers, with stars representing dog classes and points representing snake classes. While classes remain well-separated in the global space (abc), \ours{} can selectively bring disparate classes closer in their designated subspace (d) through learned feature filtering.
    }\label{fig:title}
\end{figure}

To visualize these embedding properties, we apply $t$-SNE~\citep{van2008visualizing} to the features of 10 classes from the IN1K validation set (\cref{fig:title}). The visualization reveals two key findings: (1) In the global space, \ours{} maintains clear class separation as good as the methods optimized in the global space, even through our model is optimized in subspaces; (2) When examining the subspace for specific class pairs (e.g., Garter snake-Chihuahua), our model successfully identifies shared features that bring these seemingly disparate classes closer while preserving the overall structure of the embedding space. This demonstrates our model's unique ability to learn both class-discriminative features and shared characteristics across arbitrary class pairs.

\begin{figure}
    \centering
    \includegraphics[width=0.98\linewidth]{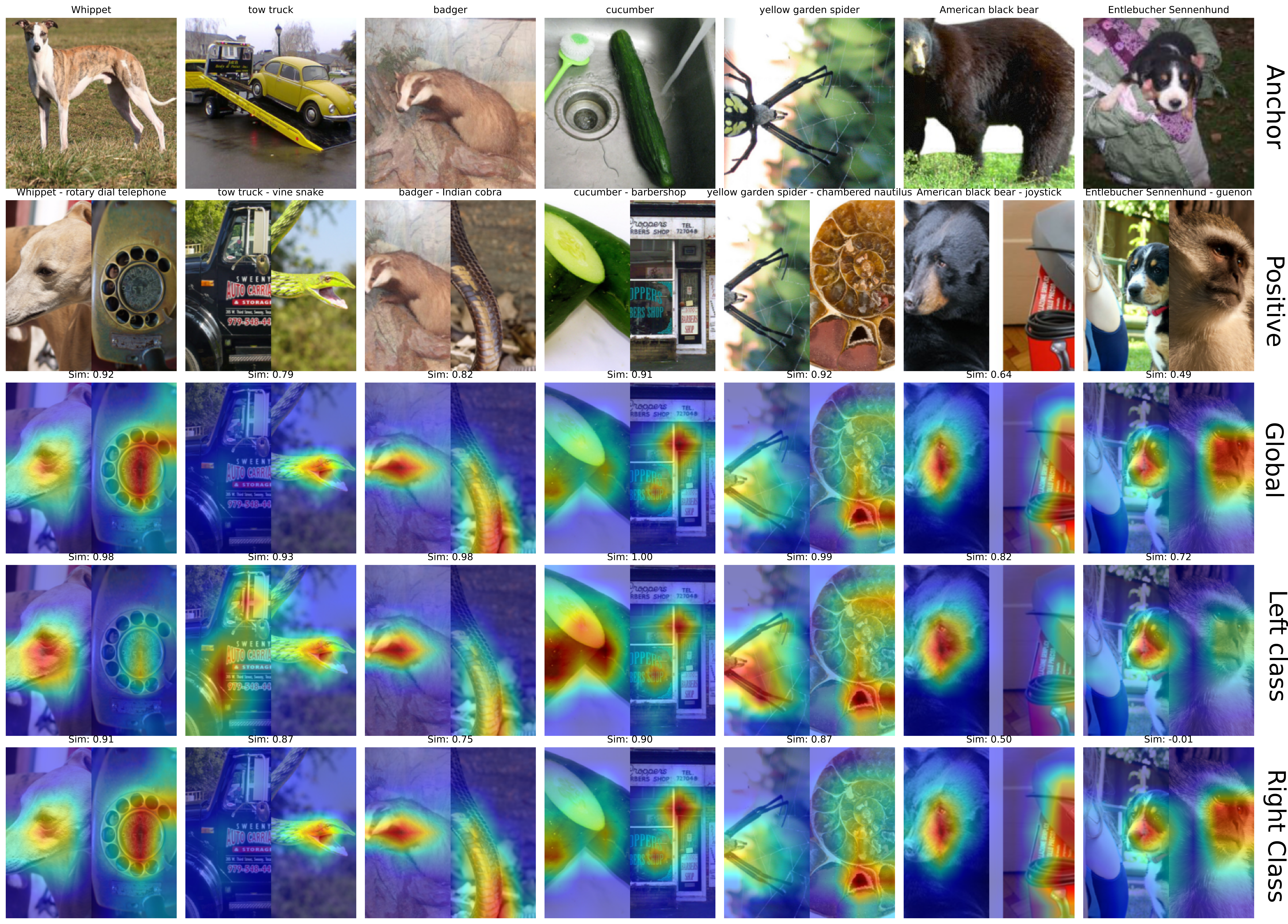}
    \caption{\textbf{Feature visualization by Grad-CAM}. Each column shows the anchor image, the positive image of a class pair $y1-y2$, and the Grad-CAM heatmaps for the features under the global space, the subspace of $y1$ and the subspace of $y2$. ``sim'' denotes the cosine similarity between the anchor and the positive.}
    \label{fig:cls_features}
\end{figure}

\paragraph{Class features.}
We utilize CLAM, a variant of Grad-CAM~\citep{zhou2016learning} (see details in \cref{sec:cam}), to understand how the feature filter selectively activates class-specific features. For each class pair $y1$-$y2$, we concatenate their images to obtain a fused representation. We then analyze how an anchor image from class $y1$ activates different regions under three scenarios: without any feature filter (global space), with gates for class $y1$, and with gates for class $y2$.
As shown in \cref{fig:cls_features}, the feature filter effectively modulates attention to class-specific features. When using gates for class $y1$, the similarity between the anchor and positive images increases as the model focuses on shared features of class $y1$. Conversely, under the subspace of class $y2$, the similarity decreases since the anchor image lacks the characteristic features of class $y2$. These visualizations demonstrate that our feature filter successfully learns to identify and selectively activate dimensions corresponding to class-specific features. This mechanism enables our framework to effectively learn from arbitrary pairs while maintaining class discrimination.

\section{Model Analysis}

\subsection{Ablation}

\begin{figure}[htb]
\begin{minipage}{.48\linewidth}    
    \centering
    \includegraphics[width=0.9\linewidth]{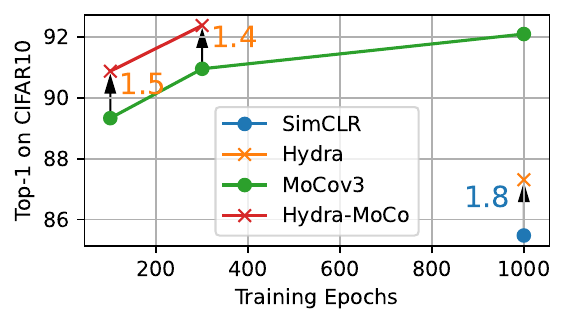}
    \caption{\textbf{Benefits of Universal Contrastive Learning.} We report KNN classification accuracy on CIFAR10. Our approach consistently improves both SimCLR and MoCov3 across different training durations, demonstrating the advantage of learning from arbitrary pairs. }
    \label{fig:efficiency}
\end{minipage} \hspace{0.03\linewidth}
\begin{minipage}{.48\linewidth}
 \includegraphics[width=0.9\linewidth]{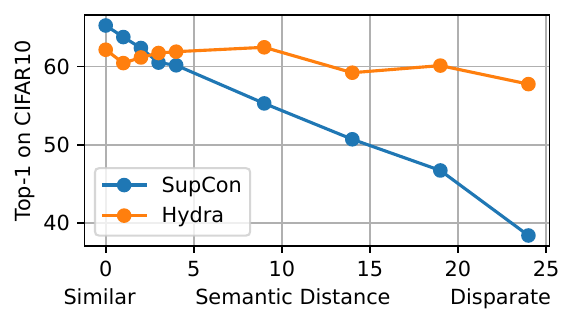}
    \caption{\textbf{Impact of semantic distance between positive pairs (N-closest classes, N=0 means same class).}  We report KNN classification accuracy on CIFAR10. While Supcon's performance drops significantly with increasing semantic distance, \ours{} maintains robust performance through adaptively learning in subspaces.}
    \label{fig:sem_dis}
\end{minipage}
\end{figure}

\paragraph{Arbitrary Pair vs. Instance Pair.}
To investigate the benefits of extending CL to arbitrary pairs, we adapt two popular instance-wise CL methods (SimCLR and MoCov3) by incorporating our feature filter. We evaluate these enhanced models on CIFAR10 using KNN classification performance across different training durations. As shown in \cref{fig:efficiency}, converting instance-wise CL to universal CL consistently improves performance. Specifically, both Hydra (enhanced SimCLR) and Hydra-MoCo (enhanced MoCov3) demonstrate superior performance compared to their base models across all training settings, highlighting the effectiveness of UCL from arbitrary pairs.

\paragraph{Range of Positive Pairs.}
To systematically evaluate the limitations of supervised contrastive learning (Supcon) and identify scenarios where our method excels, we conduct experiments on IN25, a dataset of 25 classes with well-defined semantic relationships (details in \cref{sec:data}). We systematically vary the semantic distance of positive pairs by composing them from N-closest classes, where N=0 represents pairs from the same class.
As shown in \cref{fig:sem_dis}, we compare Supcon and \ours{} on CIFAR10 transfer learning performance as we increase the semantic distance between positive pairs. Supcon's performance degrades consistently as the semantic distance increases, highlighting its limitation in learning from disparate pairs. In contrast, \ours{} maintains robust performance across different semantic distances, thanks to its ability to create appropriate subspaces for each pair.

\subsection{Hyperparameter}

We investigate the impact of each component in \ours{}. To evaluate the quality of learned representations, we apply the widely used linear prob~\citep{chen2020simple} on IN1K, KNN (K=10) on CIFAR10 as a reference of transfer learning.
\ours{} was trained on IN1K for 100 epochs. Table~\ref{tab:tune} demonstrates the effects of various hyperparameters. Our default settings for training are highlighted in \baseline{Gray}.

\begin{table*}[t]
    
\subfloat[
\textbf{Optimizer}.
\label{tab:opt}
]{
\begin{minipage}{0.22\linewidth}{\begin{center}
\tablestyle{1pt}{1.05}
\begin{tabular}{y{30}x{24}x{24}}
\textbf{opt} & IN1K & CF10  \\
\shline
\default \small{AdamW} &     70.5    & 82.9          \\
\small{LARS}  &  \textbf{70.6}   & \textbf{86.3}       \\
\end{tabular}
\end{center}}\end{minipage}
}
\hspace{0.005\linewidth}
\subfloat[
\textbf{Dimensions} of the projector.
\label{tab:dim}
]{
\begin{minipage}{0.24\linewidth}{\begin{center}
\tablestyle{1pt}{1.05}
\begin{tabular}{y{30}x{24}x{24}}
\textbf{Dim} & IN1K & CF10 \\
\shline
\default 256 & 70.5  & 82.9    \\
2048 & 70.3 & 82.9           \\
4096 & \textbf{70.6} & \textbf{84.1} 
\end{tabular}
\end{center}}\end{minipage}
}\hspace{0.005\linewidth}
\subfloat[
\textbf{Temperature} of CL. 
\label{tab:tau}
]{
\begin{minipage}{0.22\linewidth}{\begin{center}
\tablestyle{1pt}{1.05}
\begin{tabular}{y{26}x{24}x{24}x{24}}
\textbf{$\tau$} & IN1K & CF10  \\
\shline
\default 0.15 & 70.5          & 82.9          \\
0.1  & 69.8          & 82.9           \\
0.05 & \textbf{71.2} & \textbf{84.11} 
\end{tabular}
\end{center}}\end{minipage}
}\hspace{0.005\linewidth}
\subfloat[
\textbf{Coefficient} of gates.
\label{tab:lambda}
]{
\begin{minipage}{0.2\linewidth}{\begin{center}
\tablestyle{1pt}{1.05}
\begin{tabular}{y{26}x{24}x{24}x{24}}
\textbf{$\lambda$} & IN1K & CF10 \\
\shline
\default 0 & \textbf{70.5}     & 82.9    \\
1e-2  &   69.6        &   83.0    \\
1e-1  &   69.5        &   \textbf{83.2}   \\
\end{tabular}
\end{center}}\end{minipage}
}

\caption{\textbf{Tuning hyperparameters}. We report Top-1 accuracy of the linear prob (IN1K) on IN1K and KNN ($k=10$) on CIFAR10. LARS promotes transferable representations. Dimensionality improves transfer learning. $\tau=0.05$ is the best. $\lambda$ has slight effects.}\label{tab:tune}
\vspace{-0.5em}
\end{table*}

\paragraph{Normalization.}
We find that normalization at the end of the projector is critical to train \ours{}. Without normalization, the training process becomes unstable, resulting in NaN errors. Conversely, the model barely converges with Batch Normalization~\citep{IoffeS15_BN}. This is because Batch Normalization enforces each dimension to be informative (high variance), conflicting with our filter module, which selectively closes certain dimensions. Our model only functions effectively with Layer Normalization ~\citep{BaKH16_LN}.

\paragraph{Training Parameter.}
Similar to other contrastive learning methods~\citep{chen2020simple,he2020momentum}, \ours{}'s performance is influenced by the number of dimensions and the temperature parameter.

Table~\ref{tab:dim} illustrates the effect of varying the projector's dimensions. A large projection space is crucial for transfer learning. While increasing dimensions brings minimal improvement for linear probing, it significantly enhances transfer learning performance. This can be attributed to the required information capacity: 256 dimensions are sufficient to express major visual representations for dominant objects, but minor visual information beyond dominant objects requires more dimensions. 

We examined three temperature values, as shown in Table~\ref{tab:tau}. As a critical parameter for the contrastive loss, temperature significantly affects performance. While further improvements might be achieved by exploring more values, our chosen value ($\tau=0.05$) sufficiently demonstrates the value of UCL within our resource constraints.

Table~\ref{tab:opt}  compares optimizers. AdamW~\citep{LoshchilovH19_adamw} proves more efficient in training the neural network. However, LARS~\citep{you2017large} converges to a better solution with longer training (200 epochs), improving KNN performance by 3.4\%. 
By default, we use AdamW to obtain results quickly. For optimal performance, LARS~\citep{you2017large}  has been well-examined for ResNets in previous studies~\citep{he2020momentum,chen2020simple,zbontar2021barlow,Grill20_BYOL}.

\paragraph{Gate Penalty.} 
Ideally, the gates should be binary values to activate or deactivate dimensions. In practice, we utilize the Sigmoid function to generate gates, which doesn't guarantee binary outputs.  From Table~\ref{tab:lambda} shows that our objective binary gates without the penalty ($\lambda=0$). Penalizing gates is unnecessary and may even diminish linear probe performance (1\% drop). 

\todo{\paragraph{Multi subspaces.}
training with multiple subspaces}



\section{Conclusion}
    

In this paper, we have reconsidered the concept of positive pairs in contrastive learning and explored the learning framework for arbitrary pairs. Our research breaks the limitation of positive pairs and demonstrates the potential of learning common features from seemly unrelated class pairs.
The search space of positive pairs for different contrastive learning approaches. As the allowed distance for positive pairs increases, the search space expands exponentially. This expansion highlights the potential for more comprehensive and nuanced representation learning.
Our approach, \ours{}, demonstrates remarkable robustness in dealing with arbitrary pairs, even in a challenging case where positive pairs are randomly picked and most of them are disparate. The method successfully learns transferable representations under these challenging conditions, validating our core idea.

\paragraph{Limitation.} 
It is not clear what the inscrutable features are among the disparate pairs. 
When scaling our framework to a large dataset, the pair size increases as the square of the label number, which introduces more low-value pairs. A mechanism to reduce the search space is critical for applying our framework to large-scale datasets efficiently.

\newpage

\bibliographystyle{unsrtnat}
\bibliography{CL,reference}

\begin{thebibliography}{37}
\providecommand{\natexlab}[1]{#1}
\providecommand{\url}[1]{\texttt{#1}}
\expandafter\ifx\csname urlstyle\endcsname\relax
  \providecommand{\doi}[1]{doi: #1}\else
  \providecommand{\doi}{doi: \begingroup \urlstyle{rm}\Url}\fi

\bibitem[Chen et~al.(2020)Chen, Kornblith, Norouzi, and Hinton]{chen2020simple}
Ting Chen, Simon Kornblith, Mohammad Norouzi, and Geoffrey Hinton.
\newblock A simple framework for contrastive learning of visual representations.
\newblock In \emph{International conference on machine learning}, pages 1597--1607. PMLR, 2020.

\bibitem[He et~al.(2020)He, Fan, Wu, Xie, and Girshick]{he2020momentum}
Kaiming He, Haoqi Fan, Yuxin Wu, Saining Xie, and Ross Girshick.
\newblock Momentum contrast for unsupervised visual representation learning.
\newblock In \emph{Proceedings of the IEEE/CVF conference on computer vision and pattern recognition}, pages 9729--9738, 2020.

\bibitem[Khosla et~al.(2020)Khosla, Teterwak, Wang, Sarna, Tian, Isola, Maschinot, Liu, and Krishnan]{khosla2020supervised}
Prannay Khosla, Piotr Teterwak, Chen Wang, Aaron Sarna, Yonglong Tian, Phillip Isola, Aaron Maschinot, Ce~Liu, and Dilip Krishnan.
\newblock Supervised contrastive learning.
\newblock \emph{Advances in neural information processing systems}, 33:\penalty0 18661--18673, 2020.

\bibitem[Cui et~al.(2021)Cui, Zhong, Liu, Yu, and Jia]{cui2021parametric}
Jiequan Cui, Zhisheng Zhong, Shu Liu, Bei Yu, and Jiaya Jia.
\newblock Parametric contrastive learning.
\newblock In \emph{Proceedings of the IEEE/CVF international conference on computer vision}, pages 715--724, 2021.

\bibitem[Weinberger and Saul(2009)]{weinberger2009distance}
Kilian~Q Weinberger and Lawrence~K Saul.
\newblock Distance metric learning for large margin nearest neighbor classification.
\newblock \emph{Journal of machine learning research}, 10\penalty0 (2), 2009.

\bibitem[Oord et~al.(2018)Oord, Li, and Vinyals]{oord2018representation}
Aaron van~den Oord, Yazhe Li, and Oriol Vinyals.
\newblock Representation learning with contrastive predictive coding.
\newblock \emph{arXiv preprint arXiv:1807.03748}, 2018.

\bibitem[Deng et~al.(2009)Deng, Dong, Socher, Li, Li, and Fei-Fei]{deng2009imagenet}
Jia Deng, Wei Dong, Richard Socher, Li-Jia Li, Kai Li, and Li~Fei-Fei.
\newblock Imagenet: A large-scale hierarchical image database.
\newblock In \emph{2009 IEEE conference on computer vision and pattern recognition}, pages 248--255. Ieee, 2009.

\bibitem[Schroff et~al.(2015)Schroff, Kalenichenko, and Philbin]{schroff2015facenet}
Florian Schroff, Dmitry Kalenichenko, and James Philbin.
\newblock Facenet: A unified embedding for face recognition and clustering.
\newblock In \emph{Proceedings of the IEEE conference on computer vision and pattern recognition}, pages 815--823, 2015.

\bibitem[Wu et~al.(2018)Wu, Xiong, Yu, and Lin]{wu2018unsupervised}
Zhirong Wu, Yuanjun Xiong, Stella~X Yu, and Dahua Lin.
\newblock Unsupervised feature learning via non-parametric instance discrimination.
\newblock In \emph{Proceedings of the IEEE conference on computer vision and pattern recognition}, pages 3733--3742, 2018.

\bibitem[Grill et~al.(2020)Grill, Strub, Altch\'{e}, Tallec, Richemond, Buchatskaya, Doersch, Avila~Pires, Guo, Gheshlaghi~Azar, Piot, kavukcuoglu, Munos, and Valko]{Grill20_BYOL}
Jean-Bastien Grill, Florian Strub, Florent Altch\'{e}, Corentin Tallec, Pierre Richemond, Elena Buchatskaya, Carl Doersch, Bernardo Avila~Pires, Zhaohan Guo, Mohammad Gheshlaghi~Azar, Bilal Piot, koray kavukcuoglu, Remi Munos, and Michal Valko.
\newblock Bootstrap your own latent - a new approach to self-supervised learning.
\newblock In H.~Larochelle, M.~Ranzato, R.~Hadsell, M.F. Balcan, and H.~Lin, editors, \emph{Advances in Neural Information Processing Systems}, volume~33, pages 21271--21284. Curran Associates, Inc., 2020.
\newblock URL \url{https://proceedings.neurips.cc/paper_files/paper/2020/file/f3ada80d5c4ee70142b17b8192b2958e-Paper.pdf}.

\bibitem[Caron et~al.(2020)Caron, Misra, Mairal, Goyal, Bojanowski, and Joulin]{caron2020unsupervised}
Mathilde Caron, Ishan Misra, Julien Mairal, Priya Goyal, Piotr Bojanowski, and Armand Joulin.
\newblock Unsupervised learning of visual features by contrasting cluster assignments.
\newblock \emph{Advances in neural information processing systems}, 33:\penalty0 9912--9924, 2020.

\bibitem[Cui et~al.(2023)Cui, Zhong, Tian, Liu, Yu, and Jia]{cui2023generalized}
Jiequan Cui, Zhisheng Zhong, Zhuotao Tian, Shu Liu, Bei Yu, and Jiaya Jia.
\newblock Generalized parametric contrastive learning.
\newblock \emph{IEEE Transactions on Pattern Analysis and Machine Intelligence}, 2023.

\bibitem[Tsai et~al.(2021)Tsai, Li, Liu, Liao, Salakhutdinov, and Morency]{tsai2021integrating}
Yao-Hung~Hubert Tsai, Tianqin Li, Weixin Liu, Peiyuan Liao, Ruslan Salakhutdinov, and Louis-Philippe Morency.
\newblock Integrating auxiliary information in self-supervised learning.
\newblock \emph{arXiv preprint arXiv:2106.02869}, 2021.

\bibitem[Ma et~al.(2021)Ma, Tsai, Liang, Zhao, Zhang, Salakhutdinov, and Morency]{ma2021conditional}
Martin~Q Ma, Yao-Hung~Hubert Tsai, Paul~Pu Liang, Han Zhao, Kun Zhang, Ruslan Salakhutdinov, and Louis-Philippe Morency.
\newblock Conditional contrastive learning for improving fairness in self-supervised learning.
\newblock \emph{arXiv preprint arXiv:2106.02866}, 2021.

\bibitem[Cabannes et~al.(2023)Cabannes, Bottou, Lecun, and Balestriero]{Cabannes_2023_ICCV}
Vivien Cabannes, Leon Bottou, Yann Lecun, and Randall Balestriero.
\newblock Active self-supervised learning: A few low-cost relationships are all you need.
\newblock In \emph{Proceedings of the IEEE/CVF International Conference on Computer Vision (ICCV)}, pages 16274--16283, October 2023.

\bibitem[Sobal et~al.(2024)Sobal, Ibrahim, Balestriero, Cabannes, Bouchacourt, Astolfi, Cho, and LeCun]{sobal2024_Xsample}
Vlad Sobal, Mark Ibrahim, Randall Balestriero, Vivien Cabannes, Diane Bouchacourt, Pietro Astolfi, Kyunghyun Cho, and Yann LeCun.
\newblock $\mathbb{X}$-sample contrastive loss: Improving contrastive learning with sample similarity graphs.
\newblock \emph{arXiv preprint arXiv:2407.18134}, 2024.

\bibitem[Hua et~al.(2021)Hua, Wang, Xue, Ren, Wang, and Zhao]{hua2021feature}
Tianyu Hua, Wenxiao Wang, Zihui Xue, Sucheng Ren, Yue Wang, and Hang Zhao.
\newblock On feature decorrelation in self-supervised learning.
\newblock In \emph{Proceedings of the IEEE/CVF International Conference on Computer Vision}, pages 9598--9608, 2021.

\bibitem[Jing et~al.(2022)Jing, Vincent, LeCun, and Tian]{DBLP:conf/iclr/JingVLT22}
Li~Jing, Pascal Vincent, Yann LeCun, and Yuandong Tian.
\newblock Understanding dimensional collapse in contrastive self-supervised learning.
\newblock In \emph{The Tenth International Conference on Learning Representations, {ICLR} 2022, Virtual Event, April 25-29, 2022}. OpenReview.net, 2022.
\newblock URL \url{https://openreview.net/forum?id=YevsQ05DEN7}.

\bibitem[Zhang et~al.(2024)Zhang, Lan, Qu, Cheng, Feng, and Hooi]{zhang2024avoiding}
Jihai Zhang, Xiang Lan, Xiaoye Qu, Yu~Cheng, Mengling Feng, and Bryan Hooi.
\newblock Avoiding feature suppression in contrastive learning: Learning what has not been learned before.
\newblock \emph{arXiv preprint arXiv:2402.11816}, 2024.

\bibitem[Liu et~al.(2017)Liu, Wen, Yu, Li, Raj, and Song]{liu2017sphereface}
Weiyang Liu, Yandong Wen, Zhiding Yu, Ming Li, Bhiksha Raj, and Le~Song.
\newblock Sphereface: Deep hypersphere embedding for face recognition.
\newblock In \emph{Proceedings of the IEEE conference on computer vision and pattern recognition}, pages 212--220, 2017.

\bibitem[Morin and Bengio(2005)]{morin2005hierarchical}
Frederic Morin and Yoshua Bengio.
\newblock Hierarchical probabilistic neural network language model.
\newblock In \emph{International workshop on artificial intelligence and statistics}, pages 246--252. PMLR, 2005.

\bibitem[Murtagh and Contreras(2012)]{murtagh2012hierarchical}
Fionn Murtagh and Pedro Contreras.
\newblock Algorithms for hierarchical clustering: an overview.
\newblock \emph{Wiley Interdisciplinary Reviews: Data Mining and Knowledge Discovery}, 2\penalty0 (1):\penalty0 86--97, 2012.

\bibitem[He et~al.(2016)He, Zhang, Ren, and Sun]{he2016deep}
Kaiming He, Xiangyu Zhang, Shaoqing Ren, and Jian Sun.
\newblock Deep residual learning for image recognition.
\newblock In \emph{Proceedings of the IEEE conference on computer vision and pattern recognition}, pages 770--778, 2016.

\bibitem[Goyal et~al.(2021)Goyal, Duval, Reizenstein, Leavitt, Xu, Lefaudeux, Singh, Reis, Caron, Bojanowski, Joulin, and Misra]{goyal2021vissl}
Priya Goyal, Quentin Duval, Jeremy Reizenstein, Matthew Leavitt, Min Xu, Benjamin Lefaudeux, Mannat Singh, Vinicius Reis, Mathilde Caron, Piotr Bojanowski, Armand Joulin, and Ishan Misra.
\newblock Vissl.
\newblock \url{https://github.com/facebookresearch/vissl}, 2021.

\bibitem[Dosovitskiy(2020)]{dosovitskiy2020image}
Alexey Dosovitskiy.
\newblock An image is worth 16x16 words: Transformers for image recognition at scale.
\newblock \emph{arXiv preprint arXiv:2010.11929}, 2020.

\bibitem[Dehghani et~al.(2023)Dehghani, Djolonga, Mustafa, Padlewski, Heek, Gilmer, Steiner, Caron, Geirhos, Alabdulmohsin, et~al.]{dehghani2023scaling}
Mostafa Dehghani, Josip Djolonga, Basil Mustafa, Piotr Padlewski, Jonathan Heek, Justin Gilmer, Andreas~Peter Steiner, Mathilde Caron, Robert Geirhos, Ibrahim Alabdulmohsin, et~al.
\newblock Scaling vision transformers to 22 billion parameters.
\newblock In \emph{International Conference on Machine Learning}, pages 7480--7512. PMLR, 2023.

\bibitem[Van~der Maaten and Hinton(2008)]{van2008visualizing}
Laurens Van~der Maaten and Geoffrey Hinton.
\newblock Visualizing data using t-sne.
\newblock \emph{Journal of machine learning research}, 9\penalty0 (11), 2008.

\bibitem[Zhou et~al.(2016)Zhou, Khosla, Lapedriza, Oliva, and Torralba]{zhou2016learning}
Bolei Zhou, Aditya Khosla, Agata Lapedriza, Aude Oliva, and Antonio Torralba.
\newblock Learning deep features for discriminative localization.
\newblock In \emph{Proceedings of the IEEE conference on computer vision and pattern recognition}, pages 2921--2929, 2016.

\bibitem[Ioffe and Szegedy(2015)]{IoffeS15_BN}
Sergey Ioffe and Christian Szegedy.
\newblock Batch normalization: Accelerating deep network training by reducing internal covariate shift.
\newblock In Francis~R. Bach and David~M. Blei, editors, \emph{Proceedings of the 32nd International Conference on Machine Learning, {ICML} 2015, Lille, France, 6-11 July 2015}, volume~37 of \emph{{JMLR} Workshop and Conference Proceedings}, pages 448--456. JMLR.org, 2015.
\newblock URL \url{http://proceedings.mlr.press/v37/ioffe15.html}.

\bibitem[Ba et~al.(2016)Ba, Kiros, and Hinton]{BaKH16_LN}
Lei~Jimmy Ba, Jamie~Ryan Kiros, and Geoffrey~E. Hinton.
\newblock Layer normalization.
\newblock \emph{CoRR}, abs/1607.06450, 2016.
\newblock URL \url{http://arxiv.org/abs/1607.06450}.

\bibitem[Loshchilov and Hutter(2019)]{LoshchilovH19_adamw}
Ilya Loshchilov and Frank Hutter.
\newblock Decoupled weight decay regularization.
\newblock In \emph{7th International Conference on Learning Representations, {ICLR} 2019, New Orleans, LA, USA, May 6-9, 2019}. OpenReview.net, 2019.
\newblock URL \url{https://openreview.net/forum?id=Bkg6RiCqY7}.

\bibitem[You et~al.(2017)You, Gitman, and Ginsburg]{you2017large}
Yang You, Igor Gitman, and Boris Ginsburg.
\newblock Large batch training of convolutional networks.
\newblock \emph{arXiv preprint arXiv:1708.03888}, 2017.

\bibitem[Zbontar et~al.(2021)Zbontar, Jing, Misra, LeCun, and Deny]{zbontar2021barlow}
Jure Zbontar, Li~Jing, Ishan Misra, Yann LeCun, and St{\'e}phane Deny.
\newblock Barlow twins: Self-supervised learning via redundancy reduction.
\newblock In \emph{International conference on machine learning}, pages 12310--12320. PMLR, 2021.

\bibitem[Touvron et~al.(2022)Touvron, Cord, and J{\'e}gou]{touvron2022deit}
Hugo Touvron, Matthieu Cord, and Herv{\'e} J{\'e}gou.
\newblock Deit iii: Revenge of the vit.
\newblock In \emph{European conference on computer vision}, pages 516--533. Springer, 2022.

\bibitem[Wu et~al.(2024)Wu, Mo, Atito, Feng, Kittler, and Awais]{wu2024dailymae}
Jiantao Wu, Shentong Mo, Sara Atito, Zhenhua Feng, Josef Kittler, and Muhammad Awais.
\newblock Dailymae: Towards pretraining masked autoencoders in one day.
\newblock \emph{arXiv preprint arXiv:2404.00509}, 2024.

\bibitem[Selvaraju et~al.(2017)Selvaraju, Cogswell, Das, Vedantam, Parikh, and Batra]{selvaraju2017grad}
Ramprasaath~R Selvaraju, Michael Cogswell, Abhishek Das, Ramakrishna Vedantam, Devi Parikh, and Dhruv Batra.
\newblock Grad-cam: Visual explanations from deep networks via gradient-based localization.
\newblock In \emph{Proceedings of the IEEE international conference on computer vision}, pages 618--626, 2017.

\bibitem[Sammani et~al.(2023)Sammani, Joukovsky, and Deligiannis]{sammani2023visualizing}
Fawaz Sammani, Boris Joukovsky, and Nikos Deligiannis.
\newblock Visualizing and understanding contrastive learning.
\newblock \emph{IEEE Transactions on Image Processing}, 2023.

\end{thebibliography}





\appendix
\section*{Supplemental Material}

In this appendix, we present visualization on a subset of 1K classes to help understanding the behavior of our model in \cref{sec:case}. 
We present the algorithm to extend a CL method for arbitrary pairs in \ref{sec:alg}.
We discuss the potential of removing labels in \cref{sec:ssl}.
The experimental details are listed in \cref{sec:exp}.
We present CLAM to interpret learned features from CL models in ~\cref{sec:cam}. 

\section{Understanding from A Case Study}\label{sec:case}

Considering the large number and diversity of classes in IN1K, we use a case study to demonstrate how \ours{} interprets the similarity between classes. Specifically, we select 17 classes belonging to the super synset `snake.n.01', 16 classes belonging to the super synset `wading bird.n.01', and 21 classes belonging to the super synset `furniture.n.01' from the IN1K validation set. Each class contains 50 images. This case study demonstrates that \ours{} promotes

\begin{figure}[htb]
    \centering
    \includegraphics[width=0.9\linewidth]{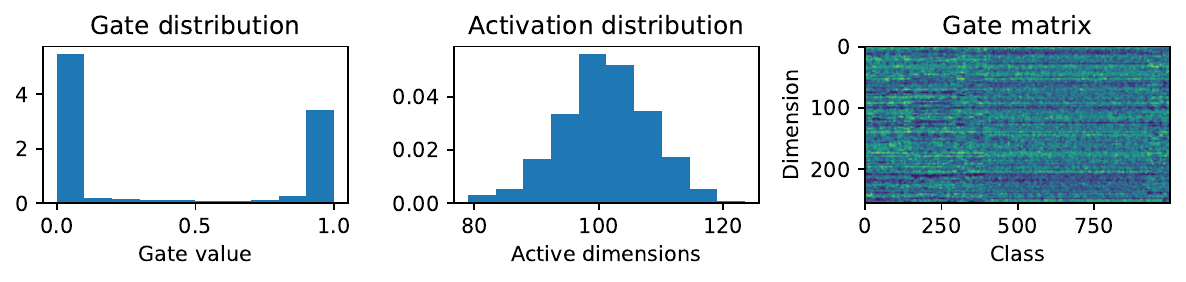}
    \caption{\textbf{Gate Visualization}. The ctivated dimensions are identified by the sum of gate values for each class, defining the size of the subspace. Gate matrix denotes the activation of one dimension for a class. The filter selects different dimensions for each class, demonstrating the creation of class-specific subspaces.}
    \label{fig:gate_dist}
\end{figure}

We summarise the gate values of \ours{} in Figure~\ref{fig:gate_dist}. Specifically, we pass 1K classes to the filter to generate 1000 vectors with 256-D gate values, which indicate the activation of the corresponding dimensions for each class. Each light point in the heatmap represents the activation status of a specific dimension for the corresponding class. From the figure, we can observe that the gates exhibit a roughly binary behavior (either activated or not), and the activation vectors for different classes are distinct. This indicates that each class creates a unique subspace within the feature space.
These experimental observations support our hypothesis that \ours{} can effectively learn to create class-specific subspaces.

\begin{figure}
    \centering
    \includegraphics[width=.9\linewidth]{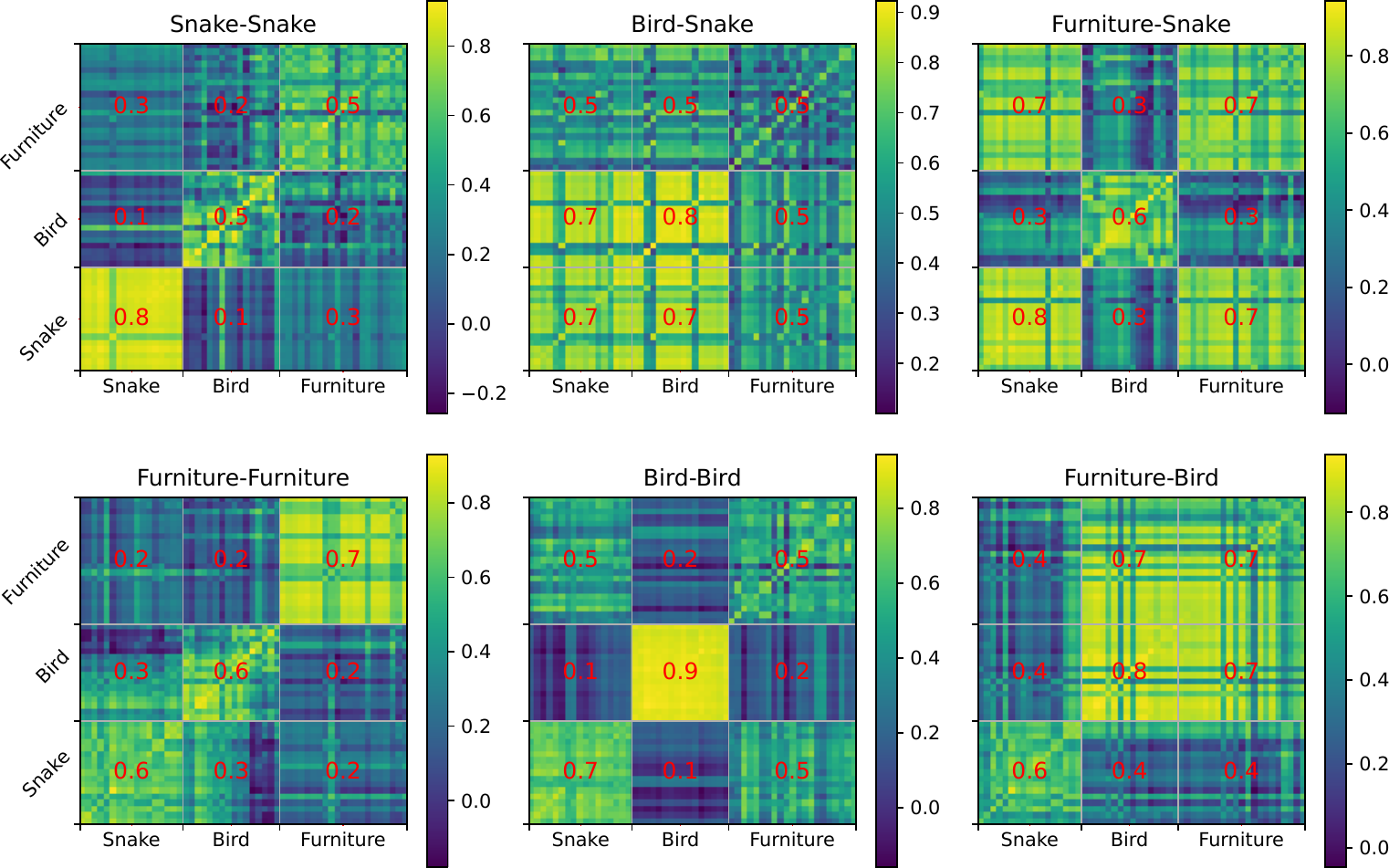}
    \caption{\textbf{Class similarity in subspaces.} The gap between intra and inter coarse-class similarity is widened in their corresponding subspaces, highlighting \ours{}'s effectiveness in creating discriminative subspaces.}
    \label{fig:rep_mani}
\end{figure}

\begin{figure}[ht]
    \begin{subfigure}{0.3\linewidth}
    \centering
        \includegraphics[width=.9\linewidth]{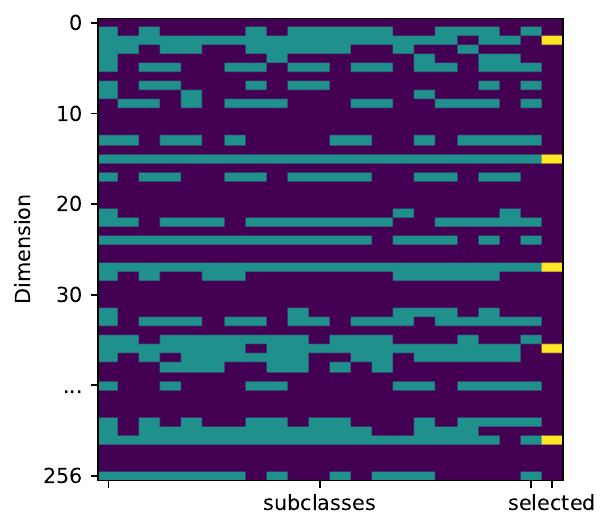}
        \subcaption[]{Gate values}
    \end{subfigure}
\caption{Example of \textbf{gate selection} for garter-lamp. Dimensions with the highest average activation are selected. }
\label{fig:gate_selection}
\end{figure}

\paragraph{Subspace of Common Features.}
We investigate how \ours{} identifies common features between similar classes within a coarse category. For the subclasses within a coarse class, the common features are dominant, i.e.  showing high similarity in the global space, as discussed earlier. 
To delve deeper, we select the top-10 dimensions with the highest activation for subclasses within each coarse class, as illustrated in Figure~\ref{fig:gate_selection}. These dimensions likely represent the most salient features for distinguishing the subclasses. We then calculate the class similarity of features in the subspace defined by these dimensions for each coarse class.
Figure~\ref{fig:rep_mani} visualizes the class similarity in the subspaces computed for three coarse classes: Snake, Bird, and Furniture. Compared to the class similarity in the global space, the discriminative information is strengthened for the corresponding subspace. Specifically, the gap between intra and inter coarse-class similarity is magnified to 0.5 for Snake in the Snake-Snake Subspace, 0.7 for Bird in the Bird-Bird Subspace, and 0.5 for Furniture in the Furniture-Furniture Subspace. 
The features in the subspaces must contain distinct information about the class pair. These features also contain visual information about other classes such that these classes are still discriminative in the unrelated subspaces. 
These results demonstrate \ours{}'s ability to create meaningful subspaces that amplify the differences between the classes, even when no common features appear to exist in the global space, like Furniture-Snake.

\begin{figure}
    \centering
    \includegraphics[width=.9\linewidth]{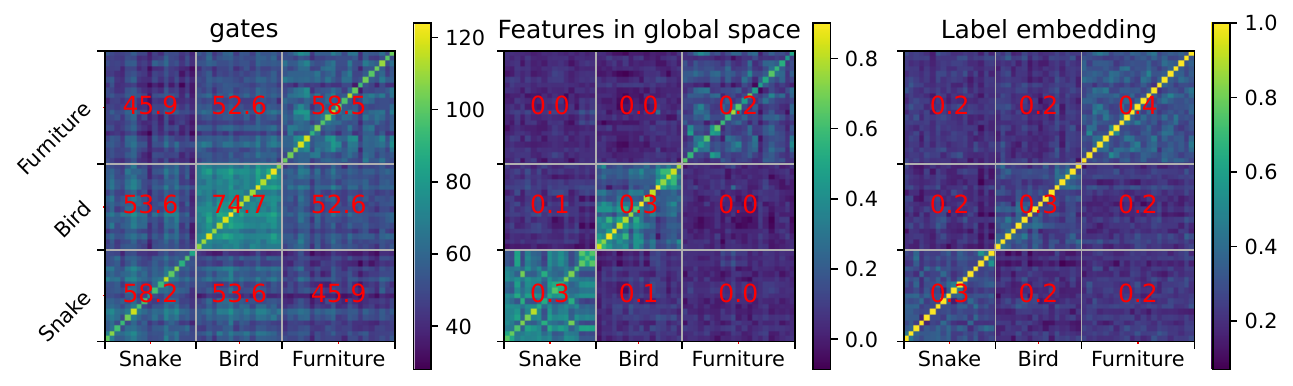}
    \caption{\textbf{Cosine similarity between classes} for gates, global features, and label vectors. Red numbers denote the average distances within each super synset region (snake, wading bird, furniture). All these measurements show that \ours{} promotes similarity between classes within the same super synset while maintaining distinctions between different super synsets.}
    \label{fig:class_dist}
\end{figure}

\paragraph{Class Similarity in Global Space.}
We investigate three similarity measurements to evaluate the ability to identify the relationships between classes: 
1) \textit{Dot product of gates}: We calculate the gates for each class and sharpen the values to 1 if above 0.5 and to 0 otherwise. This measure indicates how similar the activated dimensions are between classes, reflecting the structural similarity of their subspaces.
2) \textit{Features in the global space}: For each class pair, we use the global features to calculate the cosine similarity between all sample pairs coming from each class pair. We then use the average to denote the similarity between classes. This represents the overall visual similarity in the original feature space.
3) \textit{Class similarity in label embedding}: We obtain the class vectors through the label embedding in the feature filter, and then calculate their cosine similarity. This measure reflects the learned semantic relationships between class labels.
Figure~\ref{fig:class_dist} compares these three similarity measurements. All results indicate that \ours{} automatically discovers both semantic and visual similarities between classes, with clearer distinctions between the super synsets compared to within them. This suggests that \ours{} effectively captures the hierarchical structure of the classes, maintaining strong similarities within super synsets while preserving distinctions between them.

\section{Algorithm Summary}\label{sec:alg}
\cref{alg:main} outlines our proposed universal contrastive learning method. The algorithm begins by sampling minibatches of input data and their corresponding labels. Notably, we sample positive pairs from the mini-batch instead of augmented views. For each pair of classes $(y_{2k-1},y_{2k})$, it generates a subspace representation using the gate function $f_g$, which takes the average of the label embeddings as input. The encoder $f_E$ extracts representations for both the reference sample and its arbitrary pair. These representations are then projected into the global space using the projector $f_P$ and element-wise multiplication with the gate vector $\bm{g}_k$ to project into subspaces. Pairwise similarities are computed for all samples in the batch using cosine similarity. The contrastive loss is calculated using these similarities and a temperature parameter $\tau$. This loss encourages the model to maximize similarity between arbitrary pairs in their shared subspace while minimizing similarity with  samples from other classes. The networks $f_E$, $f_P$, $f_l$, and $f_g$ are updated to minimize this loss. After training, only the encoder network $f_E(\cdot)$ is retained for downstream tasks, discarding the gating mechanism used during training. This approach allows the model to learn transferable representations that capture common features across seemingly disparate classes.

\begin{algorithm}[!t]
\caption{\label{alg:main} Universal contrastive learning algorithm.}
\begin{algorithmic}
    \STATE \textbf{input:} batch size $N$, constant $\tau$, structure of $f_E$ and $f_P$ for encoding, structure of $f_l$ and $f_g$ for feature filter.
    \FOR{sampled minibatch $\{\bm x_k\}_{k=1}^{2N},\{\bm y_k\}_{k=1}^{2N}$}
    \STATE \textbf{for all} $k\in \{1, \ldots, N\}$ \textbf{do}
        \STATE $~~~~$draw arbitrary pairs from random sampling.
        \STATE $~~~~$\textcolor{gray}{\# the subspace for $x_{2k-1}$ and $x_{2k}$} 
        \STATE $~~~~$$\bm{g}_k = f_g((f_l(y_{2k-1})+f_l(y_{2k}))/2)$
        \STATE $~~~~$\textcolor{gray}{\# the reference sample} 
        \STATE $~~~~$$\bm h_{2k-1} = f_E({\bm x}_{2k-1})$  \textcolor{gray}{~~~~~~~~~~~~~~~~~~~~~~~~~~~~~~\# representation}
        \STATE $~~~~$$\bar{\bm z}_{2k-1} = \bm{g}_k \odot f_P({\bm h}_{2k-1})$  \textcolor{gray}{~~~~~~~~~~~~~~~~~~~~~~~~~~~~~~~~\# features in subspace}
        \STATE $~~~~$\textcolor{gray}{\# the arbitrary pair} 
        \STATE $~~~~$$\bm h_{2k} = f_E({\bm x}_{2k})$      \textcolor{gray}{~~~~~~~~~~~~~~~~~~~~~~~~~~~~~~~~~~~~~~\# representation}
        \STATE $~~~~$$\bar{\bm{z}}_{2k} = \bm{g}_k \odot f_P({\bm h}_{2k})$  \textcolor{gray}{~~~~~~~~~~~~~~~~~~~~~~~~~~~~~~~~~~~~~~~~\# features in subspace} \todo{wrong}
    \STATE \textbf{end for}
    \STATE \textbf{for all} $i\in\{1, \ldots, 2N\}$ and $j\in\{1, \dots, 2N\}$ \textbf{do}
    \STATE $~~~~$ $s_{i,j} = \bar{\bm{z}}_i^\top \bar{\bm{z}}_j / (\lVert\bar{\bm{z}}_i\rVert \lVert\bar{\bm{z}}_j\rVert)$ \textcolor{gray}{~~~~~~~~\# pairwise similarity}\\
    \STATE \textbf{end for}
    \STATE \textbf{define} $\ell(i, j)$ \textbf{as}~ $\ell(i, j) \!=\! -\log \frac{\exp(s_{i,j}/\tau)}{\sum_{k=1}^{2N} \exp(s_{i, k}/\tau)}$ \\ \STATE $\mathcal{L} = \frac{1}{2N} \sum_{k=1, y_k \neq y_i,y_k \neq y_j }^N \left[ \ell(2k\!-\!1, 2k) + \ell(2k, 2k\!-\!1)\right]$
    \STATE update networks $f_E,f_P,f_l,$ and $f_g$to minimize $\mathcal{L}$
    \ENDFOR
    \STATE \textbf{return} encoder network $f_E(\cdot)$, and throw away others
\end{algorithmic}
\end{algorithm}

\section{Self-supervised Learning}\label{sec:ssl}
One limitation of our framework is the utilization of labels. Our framework gains benefits from both supervision and contrast. It's promising to utilize a SSL model as supervision to get rid of labels.

\begin{table}[!htb]
    \centering    
\begin{tabular}{y{30}y{30}x{20}x{20}x{20}x{20}}
Model      & Label     & CF10 & CF100 & Cars & Food   \\
        \shline
SimCLR & -     & 85.5    & 61.5     & 14.8 & 51.2   \\
Supcon & IN1K & 84.8    & 60.2     & 30.9 & 53.5 \\
\ours{} & IN1K & \textbf{86.3}    & \textbf{64.6}     & \textbf{36.7} & \textbf{54.3} \\
\hline
Supcon & Pseudo &   79.5 &	55.3 &	17.0  &43.5 \\
\ours{} & Pseudo &\textbf{82.2}    & \textbf{57.7 }    & \textbf{18.2} & \textbf{46.6}  \\
    \end{tabular}
    \captionof{table}{\textbf{Labels effectiveness}. Pseudo labels are generated by SimCLR. We report top-1 accuracy of KNN (k=10).}
     \label{tab:sup_type}
\end{table}

\paragraph{Arbitrary Contrast Efficiently Utilizes Labels.}
Table~\ref{tab:sup_type} compares the KNN (k=10) performance across 4 classification tasks. The introduction of supervision significantly enhances performance on Cars and Food datasets, leveraging transferable information. However, it slightly diminishes performance on CIFAR10 and CIFAR100. Notably,  \ours{} outperforms both its unsupervised and supervised counterparts with significant improvements, particularly on the Cars dataset. These results demonstrate that common features across classes, as captured by our method, improve transferability.

\paragraph{\ours{} with Pseudo Labels}
We explore the effectiveness of \ours{} in a self-supervised setting using pseudo labels. We apply K-Means clustering to generate 1,000 pseudo labels based on features extracted from SimCLR, which yields low-quality supervision. We then train Supcon and \ours{} for 200 epochs using these pseudo labels. The results in Table~\ref{tab:sup_type} show a significant performance drop on Cars, reflecting the weak supervision provided by SimCLR. However, \ours{} consistently outperforms Supcon across all four tasks, indicating that additional information is learned from arbitrary pairs even with bad pseudo labels.

\section{Experimental Details}\label{sec:exp}

\subsection{Dataset}\label{sec:data}
\begin{figure}[htb]
    \centering
    \includegraphics[width=.9\linewidth]{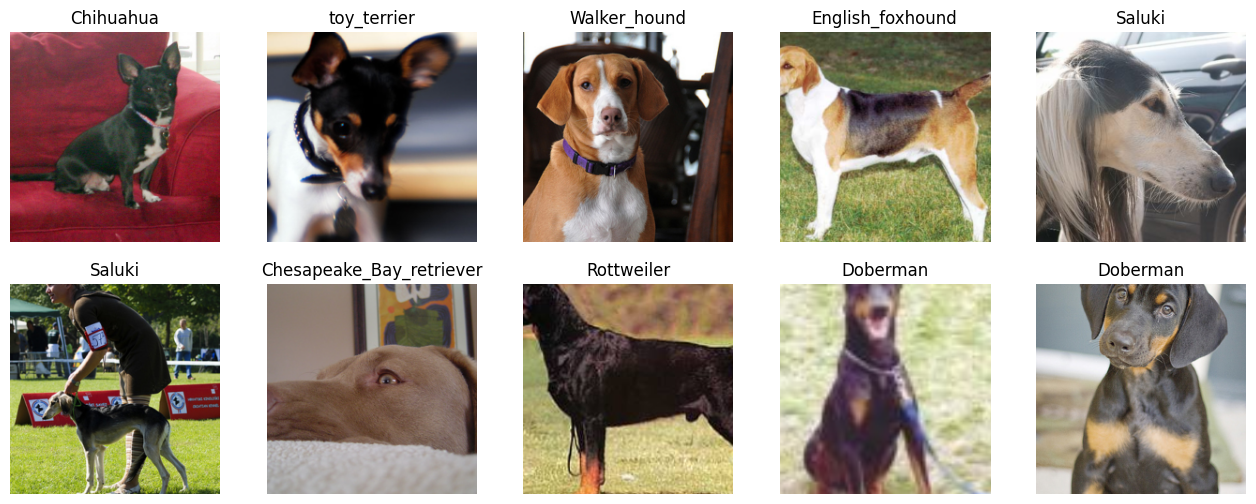}
    \caption{Samples from IN1P. }
    \label{fig:in1p_samples}
\end{figure}

\paragraph{IN1P.}
To investigate the model's ability to extract shared information among related classes, we created the IN1P dataset. This dataset comprises ten dog breeds selected from ImageNet-1K: Chihuahua, toy terrier, Walker hound, English foxhound, Saluki, Chesapeake Bay retriever, Rottweiler, Doberman, boxer, Great Dane. \cref{fig:in1p_samples} presents a sample image from each class in the IN1P dataset. Despite the variations in breed, distinct common features characteristic of dogs are evident across all samples. This carefully curated dataset allows us to examine how effectively our model can identify and leverage shared features among closely related, yet distinct, classes.

\begin{figure}
    \centering
    \includegraphics[width=.9\linewidth]{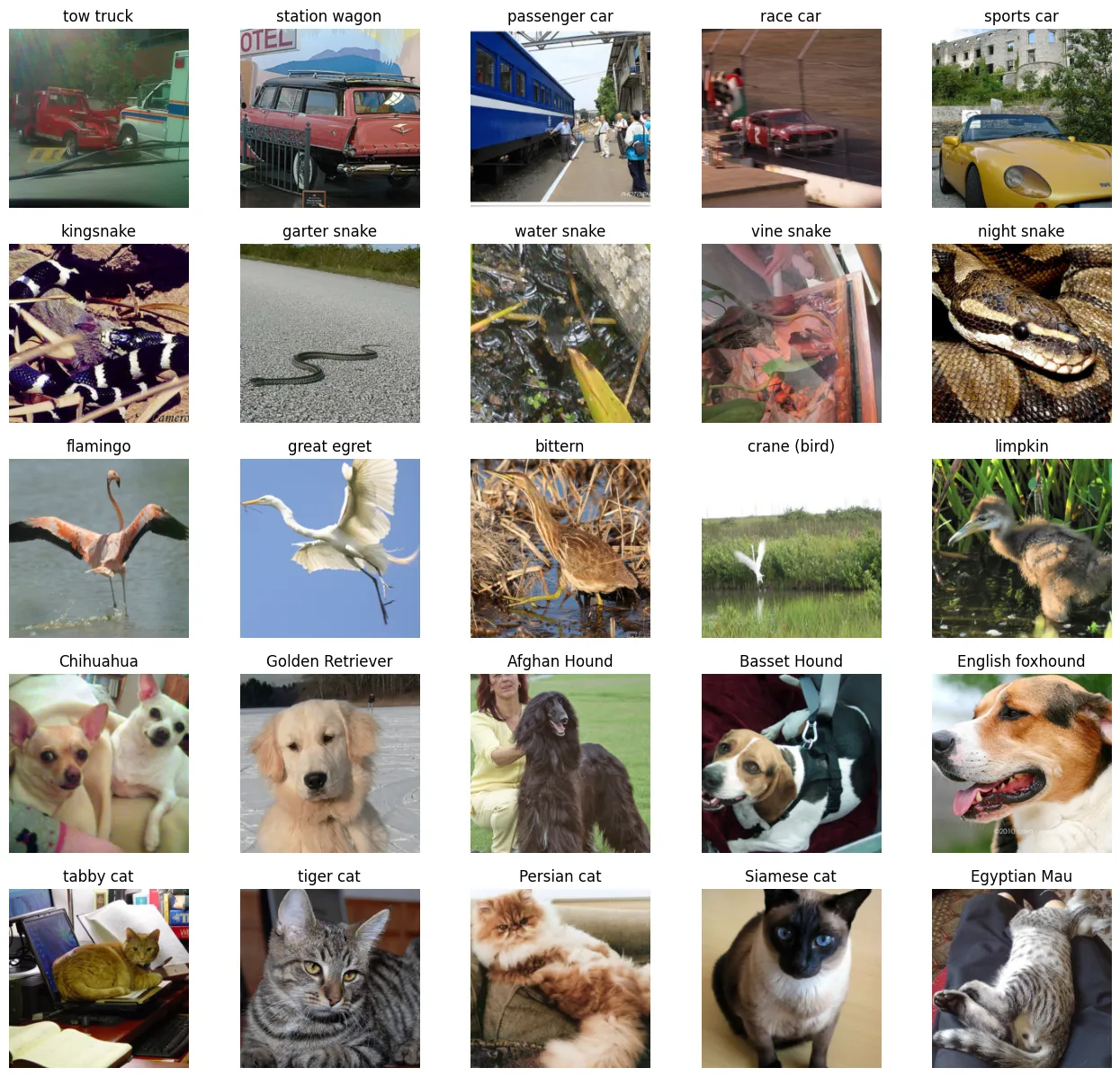}
    \caption{Samples from IN25.}
    \label{fig:in25}
\end{figure}

\begin{figure}
 \centering
\begin{subfigure}{0.45\linewidth}
    \centering
    \includegraphics[width=0.9\linewidth]{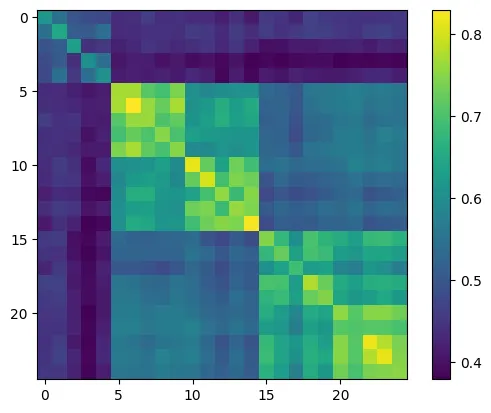}
    \subcaption{Class similarity}
    \label{fig:in25_sim_cls}    
\end{subfigure} \hspace{0.03\linewidth}
\begin{subfigure}{0.45\linewidth}
    \centering
    \includegraphics[width=0.9\linewidth]{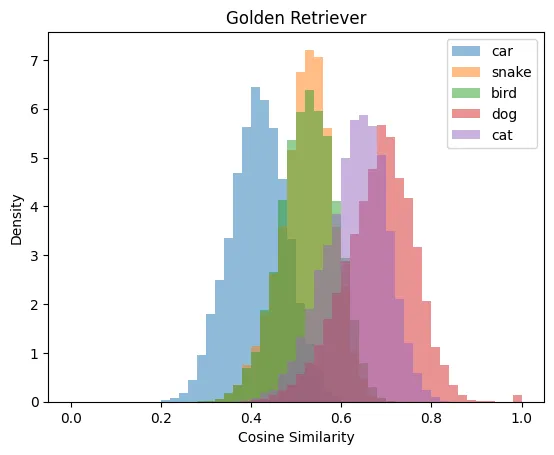}
    \subcaption{Class similarity to Golden Retriever.}
    \label{fig:in25_sim_retriever}    
\end{subfigure}
\caption{\textbf{Properties of IN25.}  IN25 demonstrates hierarchical semantic distances between classes.}
\end{figure}

\paragraph{IN25.}
We introduce IN25, a carefully curated subset of ImageNet-1K designed to study how semantic relationships affect contrastive learning. The dataset comprises 25 classes organized into 5 super-classes: cars, snakes, birds, dogs, and cats, with each super-class containing 5 sub-classes (\cref{fig:in25}). This hierarchical structure creates multiple levels of semantic relationships:
Intra-super-class: Classes within the same super-class (e.g., different dog breeds) exhibit high semantic similarity;
Inter-super-class: Classes across super-classes demonstrate varying degrees of semantic distance (e.g., dogs are semantically closer to cats than to cars).
To quantify these semantic relationships, we analyze class similarities using CLIP (ViT-B/16) embeddings. \cref{fig:in25_sim_cls} presents a heatmap of average similarities between all class pairs, revealing clear block-diagonal patterns that correspond to super-class groupings. For a more detailed view, \cref{fig:in25_sim_retriever} shows similarity distributions relative to a single class (Golden Retriever), demonstrating how semantic distances vary continuously across different super-classes.

\paragraph{Downstream Tasks.}
We use 6 downstream tasks to evaluate the transfer learning performance: 
CIFAR-10: A widely-used dataset for image classification, consisting of 60,000 32x32 color images across 10 classes, with 6,000 images per class. It includes common objects such as airplanes, automobiles, birds, and cats. The dataset is split into 50,000 training images and 10,000 test images.
CIFAR-100: Similar to CIFAR-10 but with 100 classes containing 600 images each. It maintains the same image size (32x32) and total number of images (60,000) as CIFAR-10. The classes are grouped into 20 superclasses, adding an additional layer of categorization.
STL-10: Inspired by CIFAR-10 but designed with unsupervised feature learning in mind. It contains 5,000 labeled images across 10 classes and 100,000 unlabeled images. The images are larger (96x96) and of higher quality compared to CIFAR, making it more challenging and realistic.
Stanford Cars: A fine-grained visual classification dataset containing 16,185 images of 196 classes of cars. The data is split into 8,144 training images and 8,041 testing images.
Oxford-IIIT Pet Dataset: Consists of 7,349 images of cats and dogs across 37 breeds. The dataset features 12 cat breeds and 25 dog breeds, with roughly 200 images per class. It's commonly used for tasks such as fine-grained classification and segmentation. The data is split into 3,680 training images and 3,669 testing images.
Oxford Flowers-102: A fine-grained image classification dataset comprising 102 flower categories. It contains 8,189 images in total, with each class consisting of between 40 and 258 images. The training set has only 2,040 samples. The dataset is particularly challenging due to the fine-grained nature of the categories and the large variations in scale, pose, and lighting conditions.
These tasks involve general image classification tasks and fine-grained classification tasks, providing a comprehensive evaluation of the learned representations for the image.

\subsection{Training Detail}
We conducted extensive experiments on the ImageNet-1K (IN1K) dataset. Our experimental settings are summarized in \cref{tab:train}.
For data augmentation, we use Three Augmentation~\citep{touvron2022deit}, including GaussianBlur, Gray Scaling, and Color Jitter. To accelerate the training procedure, we apply ESSL~\citep{wu2024dailymae}.
We use the efficient self-supervisede learning library to accelerate the training process~\citep{wu2024dailymae}. 

\begin{table}[]
\centering
\subfloat[
\textbf{Training parameter}. 
\label{tab:train}
]{
\begin{minipage}{0.45\linewidth}{\begin{center}
\begin{tabular}{l|cccl}
    config & \ours{}  &\ours{}  &Hydra-MoCo &\\
 arhch     &  ResNet50   &ViT-S& ResNet50 &\\
\shline
    Batch size    & 1024                 &1024                 &1024                 &\\
    epochs        & 1000 &300&1000 &\\
    Optimizer     & LARS                &AdamW&LARS                &\\
    LR            & 2e-1               &2e-3&3e-1               &\\
    LR schedule   & cosine               &cosine               &cosine               &\\
    Weight decay  & 1e-4                 &1e-2&1.5e-6                 &\\
    Warmup epochs & 40 &40 &40 &\\
    \end{tabular}
\end{center}}\end{minipage}
}

\subfloat[
\textbf{Vision Structure}. 
\label{tab:vision}
]{
\begin{minipage}{0.32\linewidth}{\begin{center}
\tablestyle{1pt}{1.05}
\begin{tabular}{l|c}
Layer & output \\
\shline
ResNet50& 2048 \\
BatchNorm1d(2048),ReLU & 2048        \\
Linear(2048,2048) & 2048        \\
BatchNorm1d(2048),ReLU & 2048        \\
Linear(2048, 256) & 256         \\ 
LayerNorm(256)    & 256        
    \end{tabular}
\end{center}}\end{minipage}
}
\hspace{0.01\linewidth}
\subfloat[
\textbf{Filter Structure}. \label{tab:filter}
]{
\begin{minipage}{0.32\linewidth}{\begin{center}
\tablestyle{1pt}{1.05}
\begin{tabular}{l|l}
Layer                   & output \\ 
\shline
Label Embedding (512)   & 512         \\
BatchNorm1d(512), ReLU  & 512         \\ 
Linear(512,1024)        & 1024        \\
BatchNorm1d(1024), ReLU & 1024        \\
Linear(1024, 256)       & 256         \\ 
Sigmoid                 & 256        
\end{tabular}
\end{center}}\end{minipage}
}
\caption{Training details.}\label{tab:recipe}
\end{table}






\section{CLAM}\label{sec:cam}

To better explain the role of learned features from CL models, we introduce CLAM based on Grad-CAM~\citep{selvaraju2017grad}, which is capable to visualize and interpret the interested region of the decision-making process. 

The basics of CL is to promote invariant representations for data augmentations, e.g., random crop. Motivated by this, we utilize multiple augmented views of the anchor image to calculate their cosine similarities to the positive image:
\begin{equation}
    \mathcal{L}_{\mathrm{CLAM}} = \mathbb{E}[\mathrm{sim}(f(\vx^+), f(\mathcal{A}(\vx)))],
\end{equation}
where $\mathcal{A}$ denotes augmentation function, including random crop, horizontal flips, and mutiplying the image by (1.0, 1.1, 0.9). 

\paragraph{Multiview.} 
The key change of CLAM is to use multiple local crops as it is commonly applied in CL. These augmented views may contain the whole object or a small portion, even background only. By increasing the number of views, we can get a stable representation for the object. Figure~\ref{fig:clam_pos} visualizes Grad-CAM when increasing the number of views.

\end{document}